\def\year{2020}\relax
\documentclass[letterpaper]{article} %
\usepackage{arxiv}  %
\usepackage{times}  %
\usepackage{helvet} %
\usepackage{courier}  %
\usepackage[hyphens]{url}  %
\usepackage{graphicx} %
\usepackage{graphicx}  %
\usepackage{amsfonts}
\usepackage{amsmath}
\usepackage{amssymb}
\usepackage{lipsum}
\usepackage{cancel}
\usepackage{censor}
\usepackage[normalem]{ulem}

\newcommand{\ra}[1]{\renewcommand{\arraystretch}{#1}}
\usepackage{booktabs}
\usepackage{multirow}
\usepackage{textcomp}

\makeatletter
\if@twocolumn
\newcommand{\whencolumns}[2]{
#2
}
\else
\newcommand{\whencolumns}[2]{
#1
}
\fi
\makeatother

\usepackage{ifthen}
\newboolean{arxiv}   
\setboolean{arxiv}{false}

\newcommand{\citet}[1]{\citeauthor{#1} \shortcite{#1}}
\newcommand{\citep}{\cite}
\newcommand{\citealp}[1]{\citeauthor{#1} \citeyear{#1}}

\title{LeDeepChef \includegraphics[scale=0.1]{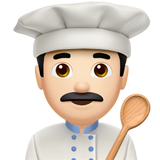} \\Deep Reinforcement Learning Agent for Families of Text-Based Games}
\author{Leonard Adolphs, Thomas Hofmann \\ Department of Computer Science\\ ETH Zurich\\ \{ladolphs, thomas.hofmann\}@inf.ethz.ch}

\begin{document}

\setboolean{arxiv}{true}
\maketitle

\begin{abstract}
While Reinforcement Learning (RL) approaches lead to significant achievements in a variety of areas in recent history, natural language tasks remained mostly unaffected, due to the compositional and combinatorial nature that makes them notoriously hard to optimize. With the emerging field of Text-Based Games (TBGs), researchers try to bridge this gap. Inspired by the success of RL algorithms on Atari games, the idea is to develop new methods in a restricted game world and then gradually move to more complex environments. Previous work in the area of TBGs has mainly focused on solving individual games. We, however, consider the task of designing an agent that not just succeeds in a single game, but performs well across a whole family of games, sharing the same theme. In this work, we present our deep RL agent---\textit{LeDeepChef}---that shows generalization capabilities to never-before-seen games of the same family with different environments and task descriptions. The agent participated in Microsoft Research\textquotesingle s \textit{First TextWorld Problems: A Language and Reinforcement Learning Challenge} and outperformed all but one competitor on the final test set. The games from the challenge all share the same theme, namely cooking in a modern house environment, but differ significantly in the arrangement of the rooms, the presented objects, and the specific goal (recipe to cook). To build an agent that achieves high scores across a whole family of games, we use an actor-critic framework and prune the action-space by using ideas from hierarchical reinforcement learning and a specialized module trained on a recipe database.
\end{abstract}

\section{Introduction}
"\textit{You are hungry! Let's cook a delicious meal. Check the cookbook in the kitchen for the recipe. Once done, enjoy your meal!}", that's the starting instruction of every game in Microsoft's \textit{First TextWorld Problems: A Language and Reinforcement Learning Challenge}; a competition that evaluates an agent on a family of unique and unseen text-based games (TBGs). While all the games share a similar theme---cooking in a modern house environment---they differ in multiple aspects like the number of rooms, connection, and arrangement of rooms, the goal of the game (i.e., different recipes), as well as actions and tools needed to succeed. TBGs are computer games where the sole modality of interaction is text. In an iterative process, the player issues commands in natural language and, in return, is presented a (partial) textual description of the environment. The player works towards goals that may or may not be specified explicitly and receives rewards upon completion. To frame it more formally, both the observation and action-space are comprised of natural language and, thus, inherit its combinatorial and compositional properties \citep{DBLP:journals/corr/abs-1806-11532}. Training an agent to succeed in such games requires overcoming several common research challenges in reinforcement learning (RL), such as partial observability, large and sparse state-, and action-space and long term credit assignment. Moreover, the agent needs several human-like abilities including understanding the environment's feedback (e.g., realize that some command had no effect on the game's state), and common sense reasoning (e.g., extracting affordance verbs to an object in the game) \citep{DBLP:journals/corr/FuldaRMW17}. \\

While TBGs reached their peak of popularity in the 1980s with games like Zork \citep{infocom}, they provide an interesting test-bed for AI agents today. Due to the dialog-like structure of the game and the goal to find a policy that maximizes the player's reward, they show great similarity to real-world tasks like question answering and open dialogue generation. Games like Zork are usually contained in a single environment that requires a variety of complex problem-solving abilities. The TextWorld framework
\citep{DBLP:journals/corr/abs-1806-11532} instead, generates a family of games with different worlds and properties but with straightforward and, most importantly, similar tasks. One can argue, that it is, therefore, more similar to human skill acquisition: once learned, a skill can easily be performed even in a slightly different environment or with new objects \citep{2019arXiv190804777Y}. Recent research has mainly focused on either learning a single TBG to high accuracy (\citealp{DBLP:journals/corr/NarasimhanKB15}; \citealp{DBLP:journals/corr/HeCHGLDO15}; \citealp{DBLP:journals/corr/abs-1812-01628}) or generalization to a completely new family of games
\citep{DBLP:journals/corr/KostkaKKR17} with only very poor performance. Microsoft's \textit{First TextWorld Problems: A Language and Reinforcement Learning Challenge} aims to cover a new research direction, that is in-between the two extremes of the single game and the general game setting. To succeed here, an agent needs to have generalization capabilities that allow it to transfer its learned \textit{cooking skills} to never-before-seen recipes in unfamiliar house environments. \\

\ifthenelse{\boolean{arxiv}}{
In this work, we present our final agent---\textit{LeDeepChef}---that achieved the high score on the (hidden) validation games and was ranked second in the overall competition. The code to train the agent, as well as an exemplary walkthrough of the game (with the agent ranking next moves), can be found on GitHub\footnote{https://github.com/leox1v/FirstTextWorldProblems}. In order to design a successful agent, we make the following contributions:}{ 
In this work, we present our final agent---\textit{LeDeepChef} ---that achieved the high score on the (hidden) validation games and was ranked second in the overall competition. The code to train the agent, as well as an exemplary walkthrough of the game (with the agent ranking next moves), can be found \sout{on GitHub}\footnote{https://github.com/\censor{leox1v/FirstTextWorldProblems}} in the Supplementary folder\footnote{Without the pre-trained weights due to submission memory contraints.}. In order to design a successful agent, we make the following contributions:}
\begin{itemize}
    \item We design an architecture that uses different parts of the context to rank a set of commands, that is trained within an actor-critic framework. Through recurrency over time-steps, we construct a model that is aware of the past context and its previous \textit{decisions}.
    \item We improve generalization to unseen environments by abstracting away standard to "high-level" commands similar to feudal learning approaches \citep{NIPS1992_714}. We show that this reduces the action-space and therefore accelerates and stabilizes the optimization procedure.
    \item We incorporate a task-specific module that predicts the missing steps to complete the task. We train it supervised on a dataset based on TextWorld recipes augmented with a list of the most common food items found in freebase to make it resilient to unseen recipes and ingredients. 
\end{itemize}
\whencolumns{
}%
{The paper is organized as follows. Section "Related Work" gives an overview of the current state of research in the field of TBGs. In section "Gameplay", we explain the TextWorld challenge in more detail and provide an exemplary walkthrough. The architecture, as well as the RL training procedure of our final agent, is described in section "Agent". Section "Command Generation" presents our command generation approach. Finally, the section "Results" compares the performance of our agent to several reasonable baselines.}

\section{Related Work}\label{sec:related_work}
Since the pioneering work of \cite{DBLP:journals/corr/MnihKSGAWR13} that combines deep neural networks with reinforcement learning techniques to successfully play Atari games, there has been an increasing interest to modify these algorithms to a variety of problems. However, due to the combinatorial and compositional property of natural language, resulting in huge action- and state-spaces, no major improvements have been made in this area. Text-based games are regarded as a good testbed for research at the intersection of RL and NLP \citep{DBLP:journals/corr/abs-1806-11532}. Even though they heavily simplify the environment---compared to, e.g., a real-world open dialogue---they present a broad spectrum of challenges for learning algorithms. 

\paragraph{Deep RL for TBGs} To solve TBGs, \cite{DBLP:journals/corr/NarasimhanKB15} developed a deep RL model that utilizes the representational power of the hidden state of Long Short-Term Memory \citep{Hochreiter:1997:LSM:1246443.1246450} to learn a Q-function. 
An adaption of this approach by \citet{DBLP:journals/corr/HeCHGLDO15}, uses two separate models to encode the context and commands individually, and then uses a pairwise interaction function between them to compute the Q-values. Since then, a variety of researchers (\citealp{DBLP:journals/corr/abs-1812-01628}; \citealp{DBLP:journals/corr/abs-1812-01628}; \citealp{2019arXiv190804777Y}; \citealp{DBLP:journals/corr/abs-1905-02265}) used some form of DQN to solve TBGs; however, we find that an advantage-actor-critic approach \citep{DBLP:journals/corr/MnihBMGLHSK16} helps to improve performance and speeds up convergence.
Using \citet{DBLP:journals/corr/NarasimhanKB15}'s LSTM-DQN or \citet{DBLP:journals/corr/HeCHGLDO15}'s adjusted DRRN on the family of games of the TextWorld challenge leads to extremely slow convergence due to the huge combinatorial action-space that arises from games with different objects and the combinatorial nature of natural language \citep{DBLP:journals/corr/abs-1812-01628}.

\paragraph{Large action-space} Text-based games can be divided by their type of input-interaction: (i) parser-based, where the agent issues commands in free form and (ii) choice-based, where the agent is presented a set of admissible commands at every turn. Assuming a fixed maximum length of the commands as well as a fixed-size vocabulary, a parser-based game is a special instance of a choice-based game with the set of all possible combinations of words in the vocabulary as the set of admissible commands. This illustrates the problem arising from combinatorial action spaces: they result in a huge set of possible options for the agent, which it cannot possibly explore in a reasonable amount of time. Hence, the major challenge is the generation of a small set of \textit{reasonable} commands for a given context. Using a supervised learning approach with a pointer-softmax model \citet{NIPS2015_5866}, \citet{DBLP:journals/corr/abs-1812-00855} as well as \citet{DBLP:journals/corr/abs-1810-05241} are able to generate admissible commands given a context for a specific TBG. A more general approach by \cite{DBLP:journals/corr/FuldaRMW17} learns to map nouns to \textit{affordant} verbs by extracting replacement vectors from word embeddings using canonical examples. \citet{NIPS2018_7615}, on the other hand, start from an over-complete set of actions and learn a binary action-elimination network by using the feedback provided by the game engine. Similarly, \citet{DBLP:journals/corr/abs-1812-01628} also prune the available actions but using a fixed scoring function on top of a graph representation of the game's state. As far as we know, our model is the first in the area of TBGs to consider grouping commands together to 'high-level' actions as a way to reduce the action-space.

\begin{figure}[t]
    \centering
    \whencolumns{%
    \includegraphics[width=0.75\columnwidth]{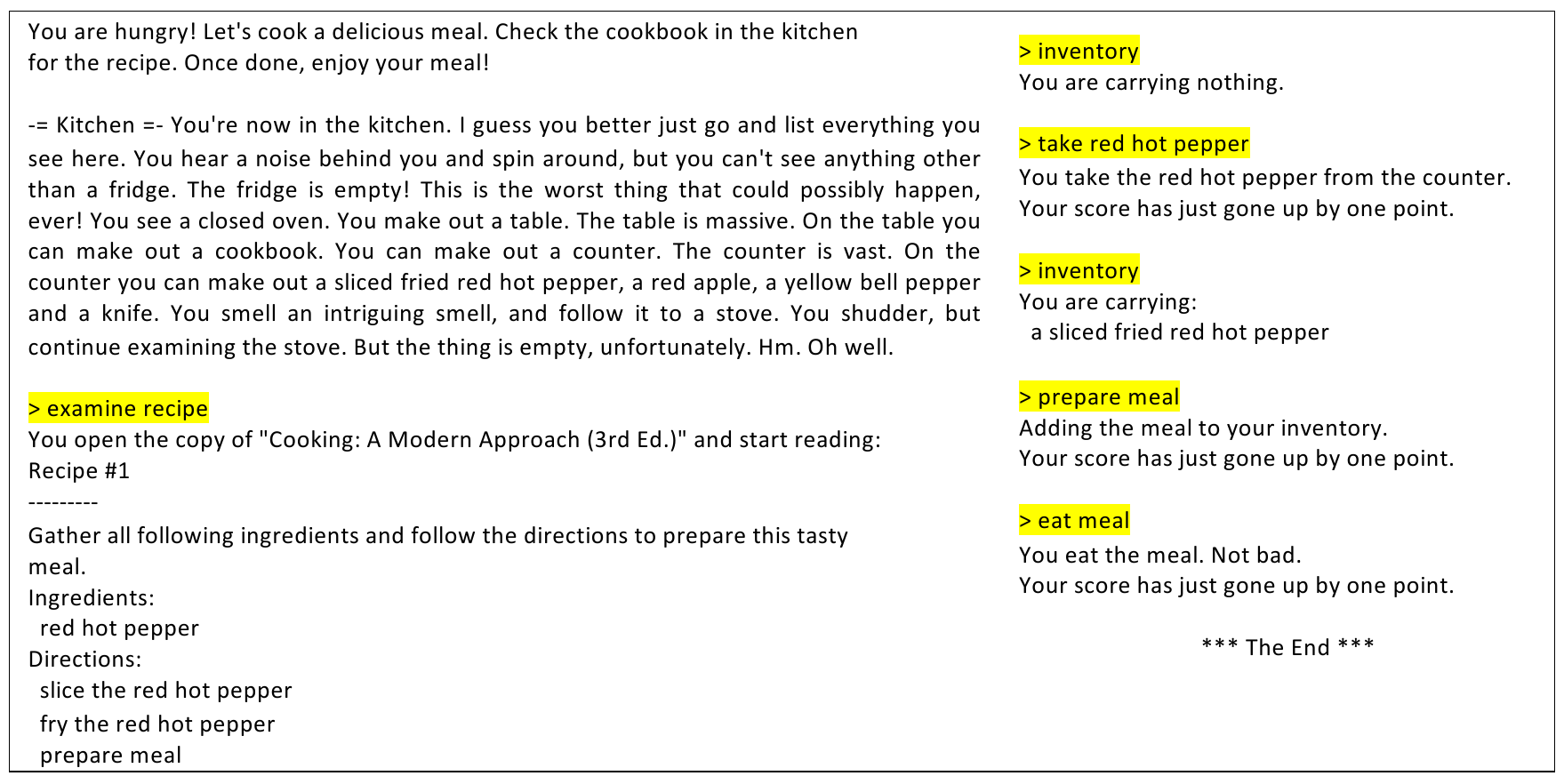}}%
    {\includegraphics[width=\columnwidth]{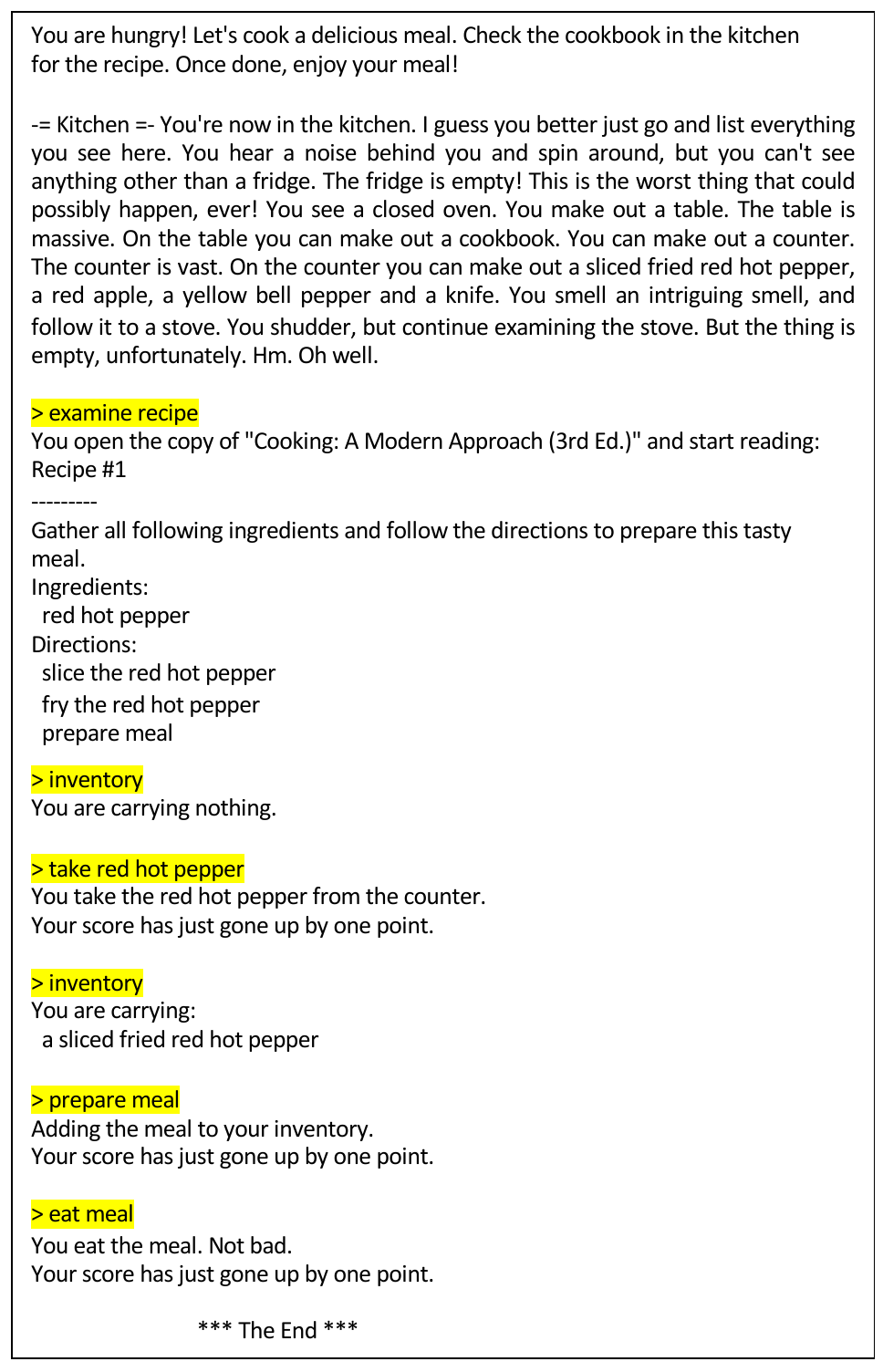}}%
    \caption{Simple game that shows the basic structure of the task. The player's commands are highlighted in yellow.}
    \label{fig:game_0}
\end{figure}

\section{Gameplay}\label{sec:gameplay}
This section provides an overview of the structure of the games in the TextWorld challenge and explains the problems an agent needs to overcome to succeed. Figure \ref{fig:game_0} shows an example of a straightforward game which helps understand the basic structure. The agent starts at a random room around the house with the instruction to find the cookbook and prepare the meal therein. The initial description of the surrounding exemplifies one of the key challenges, namely filtering the vital information from the text: sentence like \textit{you hear a noise behind you and spin around, ...} or \textit{This is the worst thing that could possibly happen, ever!} provide no useful information for the game and make it harder to understand the context. \\
Once the agent finds the room with the cookbook (in the example in Figure \ref{fig:game_0}, it is in the starting room already), the \textit{examine recipe} reveals the recipe. It consists of two parts: the ingredients, and the directions. While the ingredients part lists the items that need to be collected, the directions give information about the status they need to be in to prepare the meal. In our example, the pepper needs to be sliced and fried. Here, the agent needs to be careful, because the initial description of the surrounding states that the pepper is already sliced and fried and additional frying, for example, would lead to burning the pepper and hence losing the game. The agent, therefore, needs to remember and recognize states of ingredients mentioned in the context. With the \textit{inventory} command the agent can list all items it is currently carrying. Once all ingredients, in their correct state, are in the inventory, the agent can prepare and then eat the meal.

\section{Agent}\label{sec:agent}
We train an agent to select, at every step in the game, the most promising command (in terms of discounted future score) from a list of possible commands, given the observed context. Building a successful agent---not just for TBGs but for a wide range of sequential decision-making applications---is primarily determined by the presented set of choices at each time-step. Therefore, one of the most crucial questions is about how to generate the list of possible commands. The smaller this set is, the less time and \textit{effort} the agent wastes in its exploratory phase on "useless" strategies. To effectively reduce the size of the action-space, we use an approach inspired by hierarchical reinforcement learning, that we explain in the next section about "Command Generation". In the current section, we outline the architecture and training procedure of the agent, acting on a given set of commands.

\subsection{Model}
\begin{figure*}
    \centering
    \includegraphics[width=.8\textwidth]{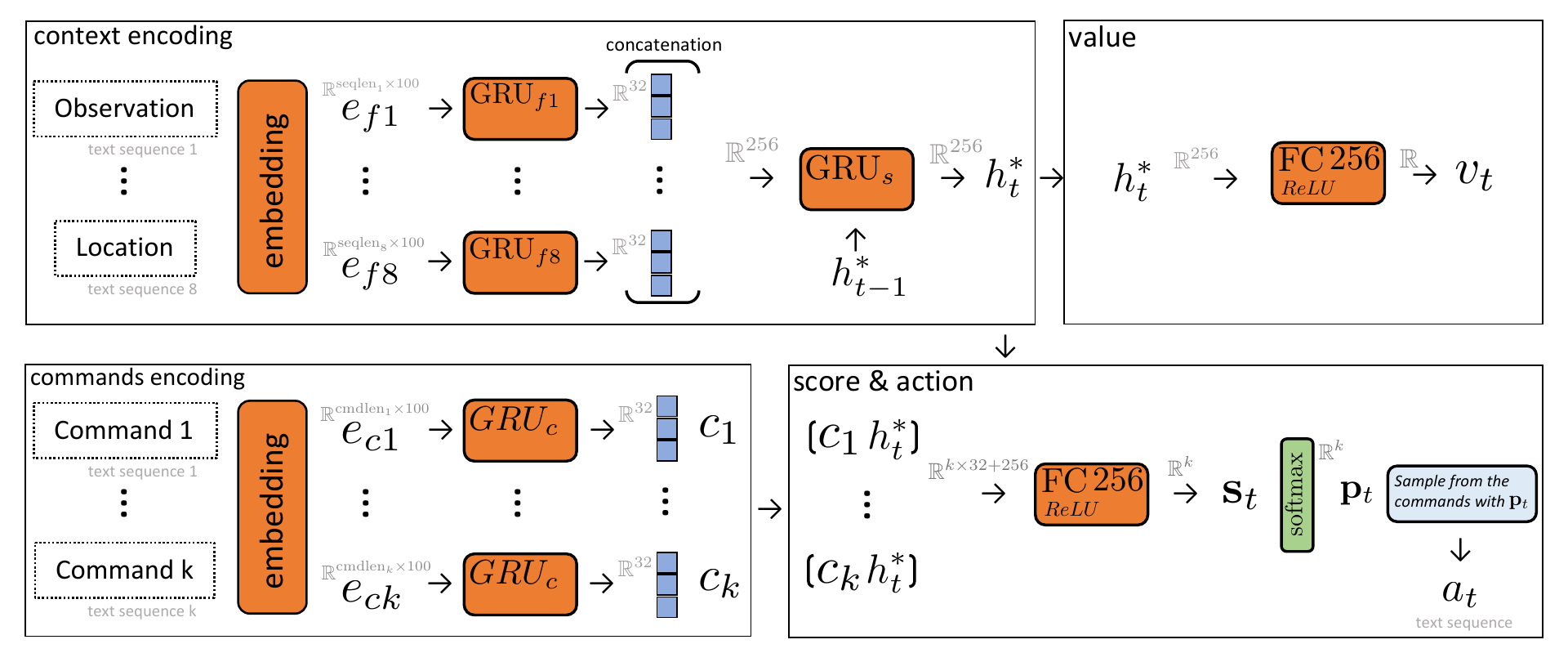}
    \caption{Illustration of the model. From a textual description of the context, together with $k$ different possible commands, it computes a categorical distribution over the commands, as well as a scalar representing the value of the current game's state.}
    \label{fig:model}
\end{figure*}
\paragraph{Model context} We build a textual context as an approximation for the (non-observable) game's state. It consists of the following text-based features:
\begin{enumerate}
    \item Observation: The response from the game engine at the current time-step. It can either be a description of what the agent sees in this room or a direct response to its last command.
    \item Missing items: The list of items that are in the recipe but not yet in the inventory. This information is constructed using the neural recipe model described in Section "Command Generation".
    \item Unnecessary items: The list of items that are in the inventory but are not needed to execute the recipe. We extract this information from the last response to the \textit{inventory} command.
    \item Description: The description of the current room. It is the output of the last \textit{look} command.
    \item Previous commands: The list of the ten previously executed commands.
    \item Required utilities: The list of kitchen appliances needed for the recipe, e.g., knife or stove. This list is a result of the prediction by the recipe model described in Section "Command Generation".
    \item Discovered locations: The list of previously visited locations.
    \item Location: The name of the current location, extracted from the last observation (if it included a location).
\end{enumerate}

The architecture of the model is shown in Figure \ref{fig:model}. It consists of four building blocks: context encoding, commands encoding, computation of the value of the current state, and the command scoring.
\paragraph{Context Encoding}
The input to the context encoding are the eight text-based features described above. Each of them is a sequence of words that we embed using a trainable 100-dimensional word embedding, initialized with pre-trained GloVe \citep{Pennington14glove:global}. This results in eight matrices of shape ($\text{seqlen}_i \times 100$ for $i=1,\dots, 8$) that are fed into eight separate bi-directional GRUs ($\text{GRU}_{fi}$). Using the last hidden vector of each GRU, we construct a fixed size encoding of size 32 for every feature input sequence. By concatenating the individual vectors, we obtain a representation for the full context with a fixed size of 256. To obtain the final context encoding $h^*$, we pass this representation into another GRU (GRU$_s$) that has its recurrency over time, i.e., it takes as hidden state the context encoding from the previous time-step \citep{DBLP:journals/corr/abs-1806-11525}.

\paragraph{Commands Encoding}
At every time-step, the model has a set with varying length $k_t$ of different possible commands to choose from. Each command is embedded using the same embedding matrix as the context, resulting in a set of $k$ matrices of size (cmdlen$_i \times 100$)\footnote{The sequence lengths of commands vary since the commands range from single words, e.g., \textit{inventory}, to short sentences, e.g., \textit{cook the red hot pepper with grill}.}. A single GRU (GRU$_c$) is used to encode the $k$ different commands individually to fixed-size representations $c_i \in \mathbb{R}^{32}$ for $i = 1,\dots, k$.

\paragraph{Value Computation}
As described in more detail in the upcoming subsection, we use an advantage-actor-critic approach to train the agent. This approach requires a \textit{critic} function that determines the value of the current state. In our model, we compute this scalar value by passing the encoded context $h^*$ through an MLP with a single hidden layer of size 256 and ReLU activations.

\paragraph{Scoring and Command Selection}
For each possible command, we compute a scalar score by feeding the concatenation of the encoded context $h^*$ and the encoded command $c_i$ for $i=1,\dots,k$ into an MLP with a single hidden layer of size 256 and ReLU activations. We obtain a score vector $\mathbf{s}_t \in \mathbb{R}^k$ that ranks the $k$ possible commands. On top of the score vector, we apply a softmax to turn it into a categorical distribution $\mathbf{p}_t$. Based on $\mathbf{p}_t$, we sample the final command from the presented set of input commands.

\subsection{Training the agent with the Actor-Critic method}\label{subsec:a2c}
We use an online actor-critic algorithm with a shared network design to optimize the agent. We compute the return $\mathcal{R}_t$ of a single time-step $t$ in the session of length $T$ by using the n-step temporal difference method \citep[ch. 7]{Sutton1998}
\begin{align}
    \mathcal{R}(s_t, a_t) = \gamma^{T-t} v(s_T) + \sum_{\tau = 0}^{T-t} \gamma^{\tau} r(s_{t+\tau}, a_{t+\tau})
\end{align}
where $\gamma$ denotes the discount factor, and $v(s_T)$ denotes the value of the state, determined by the critic network, that depends on the state $s_T$. The game-environment determines the score $r$, based on the state $s$, and the chosen action $a$.\\

From $\mathcal{R}_t$ we compute the advantage $\mathcal{A}_t$ at time-step $t$ by subtracting the state value from the critic network, i.e.
\begin{align}
    \mathcal{A}(s_t, a_t) = \mathcal{R}(s_t, a_t) - v_t(s_t).
\end{align}
While the value function from the critic $v$ captures how good a certain state is, the advantage informs us how much extra reward we obtain from action $a$ compared to the expected reward in the current state $s$.
For the sake of brevity, we will drop the indication of dependence of the state $s$ and action $a$ from now on.

\paragraph{Objective}
The full objective $\mathcal{L}$ consists of three individual terms: the policy loss, the value loss, and the entropy loss. The policy term optimizes the parameters of the actor-network while keeping the critic's weights fixed. It encourages (penalizes) the current policy if it led to a positive (negative) advantage. The policy loss is given by the following formula
\begin{align}
    \mathcal{L}_p = - \frac{1}{T}\sum_{t=1}^T \mathcal{A}_t^* \log{\boldsymbol{p}_t[a_t]}
\end{align}
where $\mathcal{A}_t^*$ is the advantage $\mathcal{A}_t$ removed from the computational graph, and $\boldsymbol{p}_t[a_t]$ is the probability of the chosen command $a_t$ determined by the actor. \\
The value term uses a mean squared error between the return $\mathcal{R}$ and the value of the critic $v_t$ to encourage them to be close, i.e.
\begin{align}
    \mathcal{L}_v = \frac{1}{2T} \sum_{t=1}^T (\mathcal{R}_t - v_t)^2 .
\end{align}
Finally, the entropy loss penalizes the actor for putting a lot of probability mass on single commands and therefore encourages exploration:
\begin{align}
    \mathcal{L}_e = -\frac{1}{T}\sum_{t=1}^T \boldsymbol{p}_t^T \log{\boldsymbol{p}_t}.
\end{align}
The final training objective is chosen as a linear combination of to three individual terms.

\section{Command Generation}\label{sec:command_generation}
One of the primary challenges in TBGs is the construction of possible---or rather \textit{reasonable}---commands in any given situation. Due to the combinatorial nature of the actions, the size of the search space is vast. Thus, brute force learning approaches are infeasible, and RL optimization is extremely difficult. We solve this problem by effectively generating only a small number of the most promising commands, as well as combining multiple actions to a single \textit{high-level} command. We find that this step of reducing the action-space is the most important to guarantee successful and stable learning of the agent. To this end, we train a helper model---called \textit{Recipe Manager}---that effectively extracts from the game's state which recipe actions still need to be performed. By comparing the state of the ingredients in the inventory with the given recipe and the description of the environment, it generates the next commands in the cooking process.

\subsection{Recipe Commands}\label{subsec:recipe_manager}
The task of this model is to determine, from the raw description of the inventory and the recipe, the following information for every listed ingredient:
\begin{itemize}
    \item Does it still need to be collected?
    \item Which cooking actions still need to be performed with it? 
\end{itemize}
Figure \ref{fig:recipe_manager} (b) in the Appendix shows an example of how the model extracts from the raw textual input the structured information needed. To achieve this, we train a model in a supervised manner with a self-constructed dataset. The dataset consists of recipes and inventories similar to those of the training games but augmented with multiple additional ingredients and adjectives to foster its generalization capabilities. Here, we query the freebase database to obtain a large selection of popular food items to make our classifier more resilient to ingredients not present in the training games.

\paragraph{Model} The input to the recipe model is the individual recipe directions and the current inventory of the agent. We do a binary classification of each direction about whether or not it needs to be performed. The necessary information about the state of the ingredient is present in the inventory. Hence, we need to map and compare each direction to it. The names of the ingredients are of varying length and can have multiple adjectives describing it, e.g., \textit{a sliced red hot pepper} or \textit{some water}. Therefore, we treat each direction and the inventory as a variable-length sequence that we encode using a GRU, after embedding it with pre-trained GloVe \citep{Pennington14glove:global}. Using pre-trained embeddings not just speeds up the convergence of the model but also helps to make it generalize across unseen ingredients, because all food-related items are close in the embedding space \citep{Pennington14glove:global}. As can be seen in Figure \ref{fig:recipe_manager} (c) in the Appendix, each of the encoded recipe directions is concatenated with the encoded inventory to serve as the input to an MLP. The network outputs a single value for each of the inputs that represent the probability of the given direction still being necessary to perform.

\paragraph{Adding recipe actions to the possible commands} 
The recipe manager adds two high-level commands to the action set. First, the \textit{take all required ingredients from here} command, grouping all the necessary 'take' commands, that can be performed in the current room. We construct this list by the intersection of needed ingredients (determined by the recipe model) and ingredients present in the current context description. Second, the \textit{drop unnecessary items} command that lists 'drop' commands for all the ingredients labeled as unnecessary from the learned recipe model. It is indeed crucial to learn to drop unwanted items because the inventory has a fixed capacity.  In addition to the abstract high-level commands, it adds all action commands---specified by the recipe model---if the specific ingredient is in the inventory and the corresponding utility in the room. Figure \ref{fig:recipe_manager} (a) in the Appendix provides an example for how the mapping from high-level to low-level commands is constructed based on the room description, the inventory, and the output from the neural recipe model.\\

\subsection{Navigation Commands}
Another crucial challenge for an agent in a TBG is to efficiently navigate through the game-world; an especially hard task when presented with unseen room configurations at test time. This process can be divided into two tasks, namely (i) understanding from the context in which direction it is possible to move, and (ii) the planning required to move through the rooms efficiently. While the latter is learned by the model as part of its policy, the challenge of extracting the movement directions from the unstructured text remains. Moreover, in the TextWorld environment, every connected room can be blocked by a closed door that the agent has to explicitly open before going into this direction. Therefore, it is necessary not only to understand in which cardinal direction to move for the next room but also to identify all closed doors in the way. For this task, we learned the \textit{Navigator} model, that is supervised trained on augmented walkthroughs to identify (i) cardinal directions that lead to connected rooms, and (ii) find all \textit{closed} doors in the current room. The model takes any room description as input and encodes the sequence with a GRU to obtain a fixed-size vector representation. This is fed into four individual MLPs that make a binary prediction on whether the corresponding cardinal direction leads to a connected room. To obtain the closed doors in the room, the hidden representation from each word of the description is fed into a shared binary MLP that predicts whether or not a particular word is \textit{part of the name of a closed door}. This approach is necessary because there can be multiple different closed doors in a room, and the name of each door can consist of multiple words, e.g., \textit{sliding patio door}. \\
The navigator adds for every \textit{detected} cardinal direction (east, south, west, north) the respective \textit{go} command to the list of possible commands. Additionally, it adds \textit{open $<$doorname$>$} for every closed door in the room's description.

\subsection{Other Commands}
Besides the commands that handle the navigation and the cooking, there are a few additional actions that are necessary to succeed in the game. Since the number of these commands is minimal, they are either added at every time-step to the set of possible commands or based on very simple rules. We provide the list of additional commands and their rules in Table \ref{tab:other_commands_rules}.
\begin{table}[h]
    \centering
    \begin{tabular}{p{0.25\linewidth}p{0.65\linewidth}}
         \textbf{Command}& \textbf{Rule}  \\
         \hline
         look, \newline inventory& Added at every step, except if they were just performed in the previous command. \\
         prepare meal & Added once the recipe manager does not output any recipe direction as missing anymore and the agent's location is the kitchen. \\
         eat meal & Added if \textit{meal} is in agent's inventory. \\
         examine cookbook & Added if the cookbook is in the room's description.
        
    \end{tabular}
    \caption{Rules for additional commands to be added to the list of possible commands.}
    \label{tab:other_commands_rules}
\end{table}

\section{Results}\label{sec:results}
First and foremost, the model was evaluated quantitatively against more than 20 competitors in Microsoft's TextWorld challenge, where it scored 1st on the (hidden) validation set and 2nd on the final test set of games. To show that our agent improves upon existing models for TBGs on never-before-seen games of the same family, we compare it against several baselines on the competition's training, validation, and test set. \\

As a metric, we always report the points per game relative to the total achievable points. A single game terminates upon successful completion of the task or when the agent fails, by either damaging an item or reaching the maximum number of a hundred steps.

\begin{figure*}
    \centering
    \begin{tabular}{p{.5\linewidth} p{.5\linewidth}}
         \includegraphics[width=.9\linewidth, height=5.5cm]{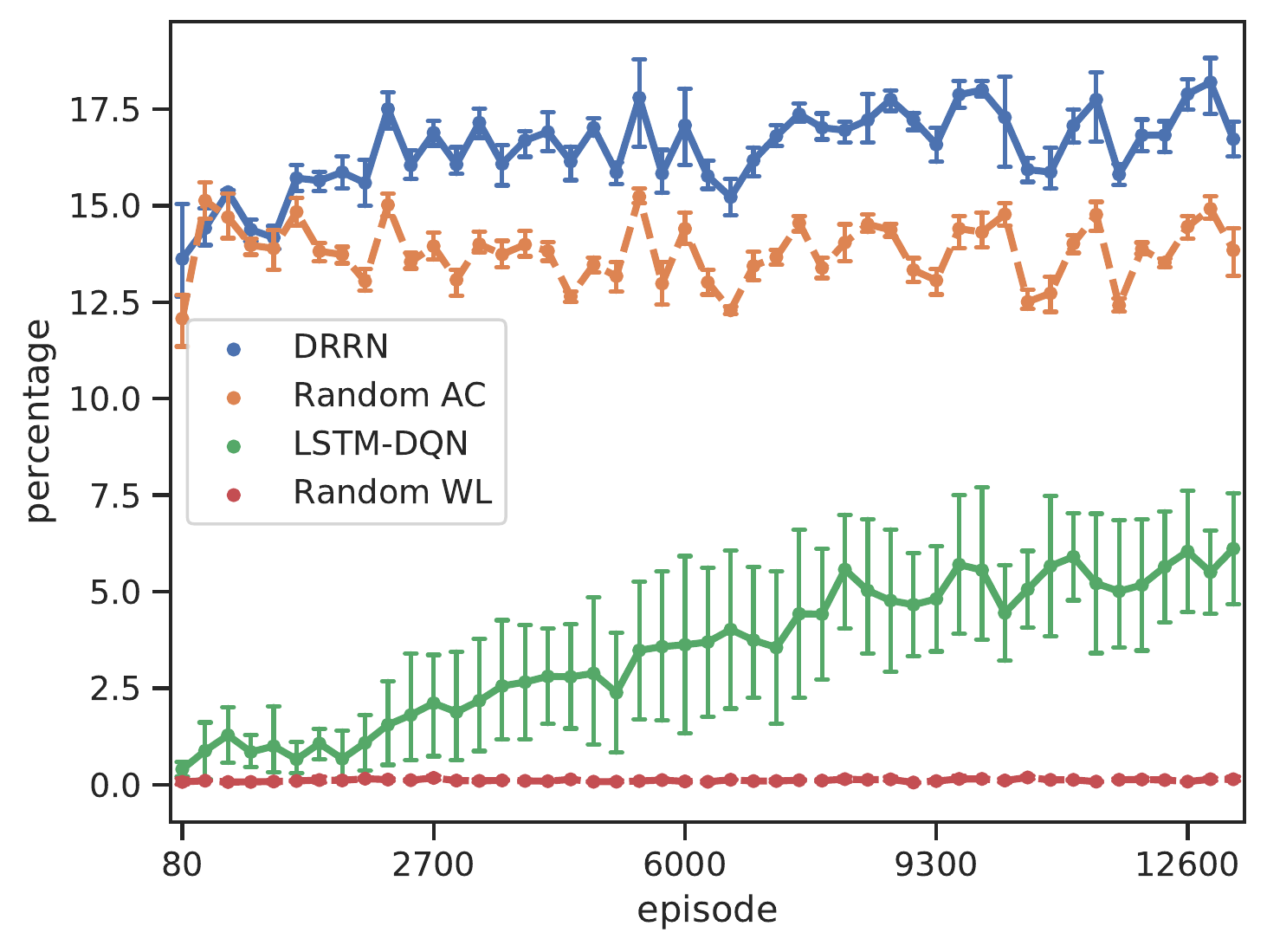} &
         \includegraphics[width=.9\linewidth, height=5.5cm]{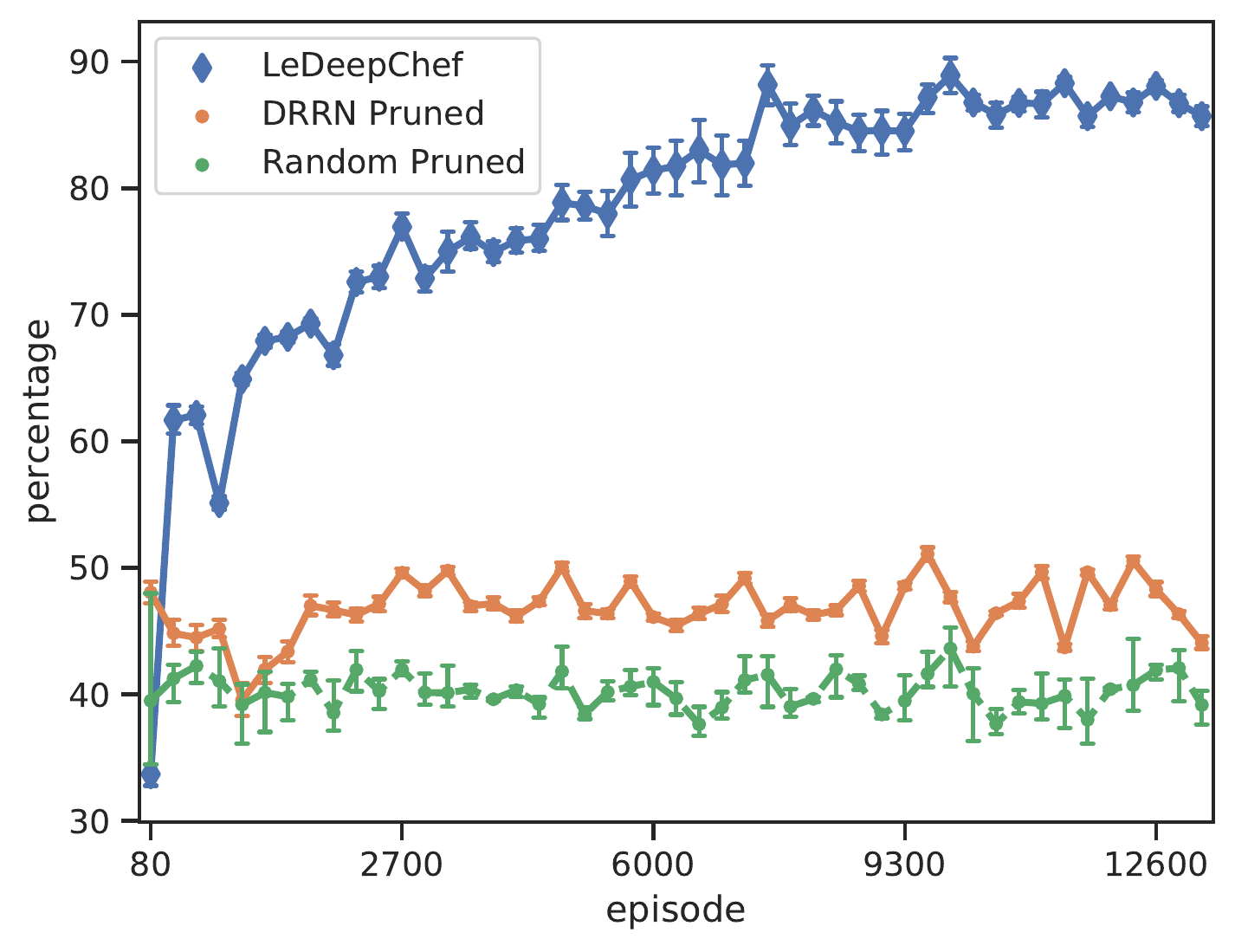} \\
         (a) Training games scoring percentage of DRRN and LSTM-DQN over three epochs. The two baselines \textit{Random AC} and \textit{Random WL} show the performance of a random agent on the admissible commands (like for DRRN) and the word-level (like LSTM-DQN), respectively. &
         (b) Training games scoring percentage of LeDeepChef compared to a DRRN model and a random baseline on the same pruned commands, generated by the recipe module.
    \end{tabular}
    \caption{Comparison of our model to several baseline models on the TextWorld challenge games, as points per game relative to total achievable points throughout the training of 3 epochs with 10 different random seeds. Each shown point is an average over the past 80 games. The model details of the baselines can be found in Table \ref{tab:model_parameters} in the Appendix.}
    \label{fig:baseline_exps}
\end{figure*}

\begin{table}
    \centering
    \ra{1.1}
    \begin{tabular}{@{}lrrrrr@{}}\toprule
    \multirow{2}[3]{*}{Method}& \multicolumn{2}{c}{valid} && \multicolumn{2}{c}{test} \\
    \cmidrule(lr){2-3} \cmidrule(lr){5-6}
     & \% & steps &  & \% & steps  \\
    \midrule

    \multirow[t]{2}{*}{Random WL}  & $0.1$ & $97.5$&  & $0.0$ & $98.9$\\
     & $\pm .04$ & $\pm .27$&  & $\pm .03$ & $\pm .02$\\
    \multirow[t]{2}{*}{LSTM-DQN}  & $2.2$ & $97.0$&  & $1.0$ & $99.3$\\
     & $\pm .00$ & $\pm .00$&  & $\pm .00$ & $\pm .00$\\
    \multirow[t]{2}{*}{Random AC}  & $11.7$ & $43.7$&  & $12.8$ & $50.1$\\
     & $\pm .59$ & $\pm 1.67$&  & $\pm .64$ & $\pm .31$\\
    \multirow[t]{2}{*}{DRRN}  & $14.0$ & $39.3$&  & $13.2$ & $50.2$\\
     & $\pm .12$ & $\pm 1.65$&  & $\pm .25$ & $\pm .05$\\
    \multirow[t]{2}{*}{Random Pruned}  & $33.5$ & $90.6$&  & $39.6$ & $95.8$\\
     & $\pm .66$ & $\pm .81$&  & $\pm .14$ & $\pm .36$\\
    \multirow[t]{2}{*}{DRRN Pruned}  & $34.3$ & $89.8$&  & $44.1$ & $92.2$\\
     & $\pm .31$ & $\pm .41$&  & $\pm 2.01$ & $\pm 1.80$\\
     \hline
    \multirow[t]{2}{*}{\citet{2019arXiv190804777Y}}  &$58$\footnotemark & $30$ &  & - &- \\
     & -& -&  & -& -\\
    \multirow[t]{2}{*}{LeDeepChef \includegraphics[scale=0.07]{src/male-cook.png}}  & $\boldsymbol{74.4}$ & $\boldsymbol{24.1}$&  & $\boldsymbol{69.3}$ & $\boldsymbol{43.9}$\\
     & $\pm .18$ & $\pm .23$&  & $\pm .20$ & $\pm .19$\\
    
    \bottomrule

    \end{tabular}
    \caption{Results on the unseen set of validation and test games from the TextWorld Challenge. We report the mean and standard deviation over ten runs with different random seeds of each best performing model on the training set.}
    \label{tab:testset_result}
\end{table}

\footnotetext{Result on their own validation set, which is hold-out data from the official training set of games. However, the dynamics and difficulty of both sets of games are comparable.}

\paragraph{Baseline} Figure \ref{fig:baseline_exps} (a) demonstrates that standard baselines for TBGs are not able to learn generalization capabilities to sufficiently solve a whole family of games. Both, LSTM-DQN \citep{DBLP:journals/corr/NarasimhanKB15} and DRRN \citep{DBLP:journals/corr/HeCHGLDO15}, do not exceed the 20\% mark of points per game relative to total achievable points\footnote{20\% would be equivalent to 12th place in the competition if the admissible commands are known at every step, as for DRRN (= handicap 5), or 9th place if not (= LSTM-DQN).} during 3 epochs of training. The input to both of these models is the concatenated game's state, consisting of the room's description, the agent's inventory, the recipe, the feedback from the last command, and the set of previously issued commands. The main difference between DRRN and LSTM-DQN is that the former ranks the provided admissible commands, while the latter ranks (pre-selected) verbs and objects, from which a command is formed then. Due to the combinatorial nature of possible commands from the LSTM-DQN, the effective action-space is significantly larger than for DRRN. Thus, a random agent on this \textit{word-level} task---\textit{Random WL}---performs much worse than an agent that selects randomly from the admissible commands, \textit{Random AC}. Both DRRN and LSTM-DQN significantly outperform their random counterpart over the course of the training but are not able to learn to solve the games to a sufficient degree. The big scoring difference between the two random agents underlines the importance of effective action-space reduction.

\paragraph{Comparison on pruned commands} In a second experiment, we use the same DRRN architecture as before, but with a pruned version of the admissible commands to exactly match the commands presented to our model; though, without the grouping to high-level actions. As we see in Figure \ref{fig:baseline_exps} (b), the reduced set of possible commands massively improves both the random and the DRRN model to up to 50\%\footnote{50\% is equivalent to the 5th place in the competition.}. However, the DRRN model is still not capable of improving a lot upon the random model and---as before---does not show a steady upward trend throughout the training procedure. Our model, on the other hand, improves its percentage significantly over the training iterations to its final score of around $87\%$. We believe that the advantage of our model over this specific baseline is mainly due to (i) the grouped high-level commands that let the agent learn a strategy more efficiently in an abstract space, (ii) the improvements in the neural architecture that acts on a more sophisticated version of the input features, and (iii) the superiority of the actor-critic over the DQN approach. 

\paragraph{Comparison on Test Set} Table \ref{tab:testset_result} shows the quantitative results of different models on the (unseen) validation set, as well as the final test set of Microsoft's TextWorld challenge. As expected, our model generalizes best to the unseen games with a mean percentage of $74.4$ and $69.3$ for the respective sets of games. The standard baselines are not able to exceed the $15\%$ mark, indicating that they are not suitable to be applied "out-of-the-box" on the specific task of solving families of TBGs. A recent model by \citet{2019arXiv190804777Y}, designed explicitly for the TextWorld environment, uses a curriculum learning approach to train a DQN model and achieves $58\%$ on their validation set (hold-out data from the challenge's training set).

\section{Conclusion}\label{sec:conclusion}
In this work, we presented how to build a deep RL agent that not only performs well on a single TBG but generalizes to never-before-seen games of the same family. To achieve this result, we designed a model that effectively ranks a set of commands based on the context and context-derived features. By incorporating ideas from hierarchical RL, we significantly reduced the size of the action-space and were able to train the agent through an actor-critic approach. Additionally, we showed how to make the agent more resilient against never-before-seen recipes and ingredients by training with data augmented by a food-item database. The performance of our final agent on the unseen games of the FirstTextWorld challenge is substantially higher than any standard baseline. Moreover, it achieved the highest score, among more than 20 competitors, on the (unseen) validation set and beat all but one agent on the final test set.

\ifthenelse{\boolean{arxiv}}{\newpage}{}

\bibliographystyle{aaai}
\bibliography{ftwp_arxiv}

\begin{thebibliography}{}

\bibitem[\protect\citeauthoryear{Ammanabrolu and
  Riedl}{2018}]{DBLP:journals/corr/abs-1812-01628}
Ammanabrolu, P., and Riedl, M.~O.
\newblock 2018.
\newblock Playing text-adventure games with graph-based deep reinforcement
  learning.
\newblock {\em CoRR} abs/1812.01628.

\bibitem[\protect\citeauthoryear{C{\^{o}}t{\'{e}} \bgroup et al\mbox.\egroup
  }{2018}]{DBLP:journals/corr/abs-1806-11532}
C{\^{o}}t{\'{e}}, M.; K{\'{a}}d{\'{a}}r, {\'{A}}.; Yuan, X.; Kybartas, B.;
  Barnes, T.; Fine, E.; Moore, J.; Hausknecht, M.~J.; Asri, L.~E.; Adada, M.;
  Tay, W.; and Trischler, A.
\newblock 2018.
\newblock Textworld: {A} learning environment for text-based games.
\newblock {\em CoRR} abs/1806.11532.

\bibitem[\protect\citeauthoryear{Dayan and Hinton}{1993}]{NIPS1992_714}
Dayan, P., and Hinton, G.~E.
\newblock 1993.
\newblock Feudal reinforcement learning.
\newblock In Hanson, S.~J.; Cowan, J.~D.; and Giles, C.~L., eds., {\em Advances
  in Neural Information Processing Systems 5}. Morgan-Kaufmann.
\newblock  271--278.

\bibitem[\protect\citeauthoryear{Fulda \bgroup et al\mbox.\egroup
  }{2017}]{DBLP:journals/corr/FuldaRMW17}
Fulda, N.; Ricks, D.; Murdoch, B.; and Wingate, D.
\newblock 2017.
\newblock What can you do with a rock? affordance extraction via word
  embeddings.
\newblock {\em CoRR} abs/1703.03429.

\bibitem[\protect\citeauthoryear{He \bgroup et al\mbox.\egroup
  }{2015}]{DBLP:journals/corr/HeCHGLDO15}
He, J.; Chen, J.; He, X.; Gao, J.; Li, L.; Deng, L.; and Ostendorf, M.
\newblock 2015.
\newblock Deep reinforcement learning with an unbounded action space.
\newblock {\em CoRR} abs/1511.04636.

\bibitem[\protect\citeauthoryear{Hochreiter and
  Schmidhuber}{1997}]{Hochreiter:1997:LSM:1246443.1246450}
Hochreiter, S., and Schmidhuber, J.
\newblock 1997.
\newblock Long short-term memory.
\newblock {\em Neural Comput.} 9(8):1735--1780.

\bibitem[\protect\citeauthoryear{Infocom}{1980}]{infocom}
Infocom.
\newblock 1980.
\newblock Zork i.

\bibitem[\protect\citeauthoryear{Kostka \bgroup et al\mbox.\egroup
  }{2017}]{DBLP:journals/corr/KostkaKKR17}
Kostka, B.; Kwiecien, J.; Kowalski, J.; and Rychlikowski, P.
\newblock 2017.
\newblock Text-based adventures of the golovin {AI} agent.
\newblock {\em CoRR} abs/1705.05637.

\bibitem[\protect\citeauthoryear{Mnih \bgroup et al\mbox.\egroup
  }{2013}]{DBLP:journals/corr/MnihKSGAWR13}
Mnih, V.; Kavukcuoglu, K.; Silver, D.; Graves, A.; Antonoglou, I.; Wierstra,
  D.; and Riedmiller, M.~A.
\newblock 2013.
\newblock Playing atari with deep reinforcement learning.
\newblock {\em CoRR} abs/1312.5602.

\bibitem[\protect\citeauthoryear{Mnih \bgroup et al\mbox.\egroup
  }{2016}]{DBLP:journals/corr/MnihBMGLHSK16}
Mnih, V.; Badia, A.~P.; Mirza, M.; Graves, A.; Lillicrap, T.~P.; Harley, T.;
  Silver, D.; and Kavukcuoglu, K.
\newblock 2016.
\newblock Asynchronous methods for deep reinforcement learning.
\newblock {\em CoRR} abs/1602.01783.

\bibitem[\protect\citeauthoryear{Narasimhan, Kulkarni, and
  Barzilay}{2015}]{DBLP:journals/corr/NarasimhanKB15}
Narasimhan, K.; Kulkarni, T.~D.; and Barzilay, R.
\newblock 2015.
\newblock Language understanding for text-based games using deep reinforcement
  learning.
\newblock {\em CoRR} abs/1506.08941.

\bibitem[\protect\citeauthoryear{Pennington, Socher, and
  Manning}{2014}]{Pennington14glove:global}
Pennington, J.; Socher, R.; and Manning, C.~D.
\newblock 2014.
\newblock Glove: Global vectors for word representation.
\newblock In {\em In EMNLP}.

\bibitem[\protect\citeauthoryear{Sutton and Barto}{2018}]{Sutton1998}
Sutton, R.~S., and Barto, A.~G.
\newblock 2018.
\newblock {\em Reinforcement Learning: An Introduction}.
\newblock The MIT Press, second edition.

\bibitem[\protect\citeauthoryear{Tao \bgroup et al\mbox.\egroup
  }{2018}]{DBLP:journals/corr/abs-1812-00855}
Tao, R.~Y.; C{\^{o}}t{\'{e}}, M.; Yuan, X.; and Asri, L.~E.
\newblock 2018.
\newblock Towards solving text-based games by producing adaptive action spaces.
\newblock {\em CoRR} abs/1812.00855.

\bibitem[\protect\citeauthoryear{Vinyals, Fortunato, and
  Jaitly}{2015}]{NIPS2015_5866}
Vinyals, O.; Fortunato, M.; and Jaitly, N.
\newblock 2015.
\newblock Pointer networks.
\newblock In Cortes, C.; Lawrence, N.~D.; Lee, D.~D.; Sugiyama, M.; and
  Garnett, R., eds., {\em Advances in Neural Information Processing Systems
  28}. Curran Associates, Inc.
\newblock  2692--2700.

\bibitem[\protect\citeauthoryear{Yin and
  May}{2019a}]{DBLP:journals/corr/abs-1905-02265}
Yin, X., and May, J.
\newblock 2019a.
\newblock Comprehensible context-driven text game playing.
\newblock {\em CoRR} abs/1905.02265.

\bibitem[\protect\citeauthoryear{{Yin} and {May}}{2019b}]{2019arXiv190804777Y}
{Yin}, X., and {May}, J.
\newblock 2019b.
\newblock {Learn How to Cook a New Recipe in a New House: Using Map
  Familiarization, Curriculum Learning, and Common Sense to Learn Families of
  Text-Based Adventure Games}.
\newblock {\em arXiv e-prints}  arXiv:1908.04777.

\bibitem[\protect\citeauthoryear{Yuan \bgroup et al\mbox.\egroup
  }{2018a}]{DBLP:journals/corr/abs-1806-11525}
Yuan, X.; C{\^{o}}t{\'{e}}, M.; Sordoni, A.; Laroche, R.; des Combes, R.~T.;
  Hausknecht, M.~J.; and Trischler, A.
\newblock 2018a.
\newblock Counting to explore and generalize in text-based games.
\newblock {\em CoRR} abs/1806.11525.

\bibitem[\protect\citeauthoryear{Yuan \bgroup et al\mbox.\egroup
  }{2018b}]{DBLP:journals/corr/abs-1810-05241}
Yuan, X.; Wang, T.; Meng, R.; Thaker, K.; He, D.; and Trischler, A.
\newblock 2018b.
\newblock Generating diverse numbers of diverse keyphrases.
\newblock {\em CoRR} abs/1810.05241.

\bibitem[\protect\citeauthoryear{Zahavy \bgroup et al\mbox.\egroup
  }{2018}]{NIPS2018_7615}
Zahavy, T.; Haroush, M.; Merlis, N.; Mankowitz, D.~J.; and Mannor, S.
\newblock 2018.
\newblock Learn what not to learn: Action elimination with deep reinforcement
  learning.
\newblock In Bengio, S.; Wallach, H.; Larochelle, H.; Grauman, K.;
  Cesa-Bianchi, N.; and Garnett, R., eds., {\em Advances in Neural Information
  Processing Systems 31}. Curran Associates, Inc.
\newblock  3562--3573.

\end{thebibliography}

\newpage
\appendix

\onecolumn
\section{Illustrations of the Recipe Manager}\label{app:Recipe_manager}

\begin{figure}[h]
    \centering
    \begin{tabular}{@{}p{.4\linewidth} p{.4\linewidth}@{}}
        \multicolumn{1}{c}{\includegraphics[width=.25\linewidth]{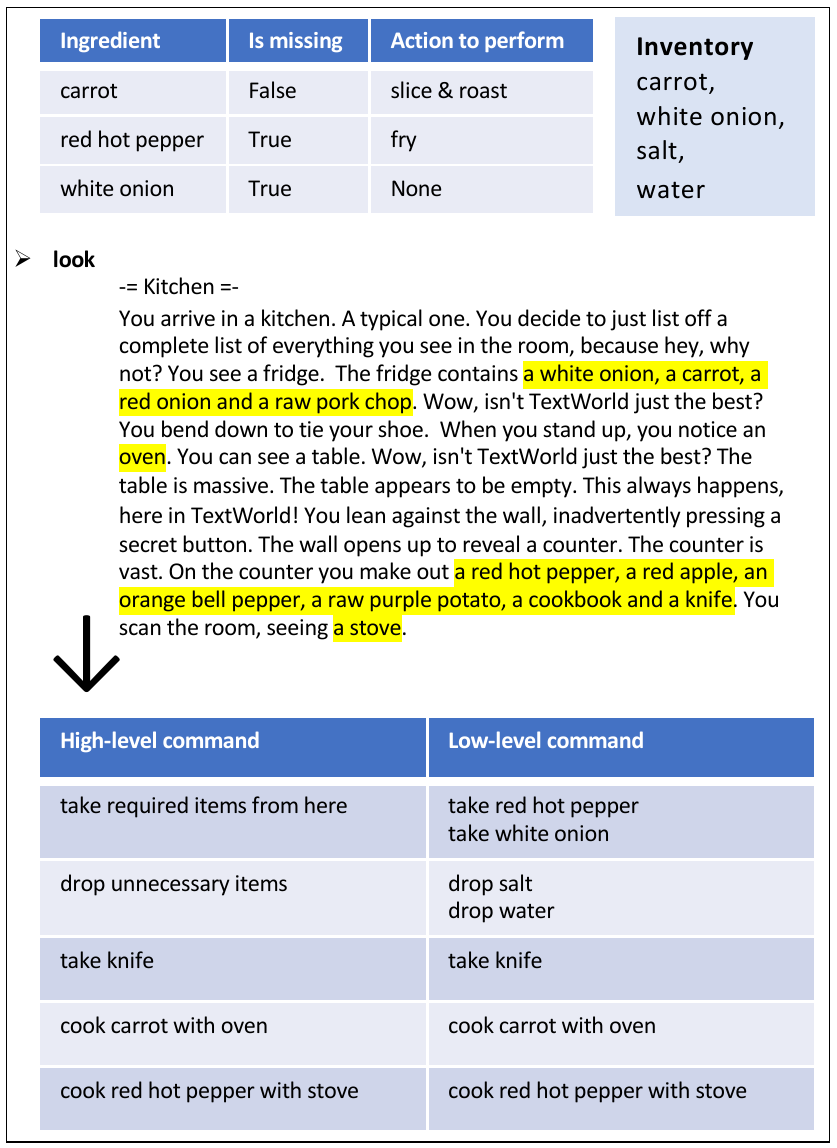}} &
        \multicolumn{1}{c}{\includegraphics[width=.25\linewidth]{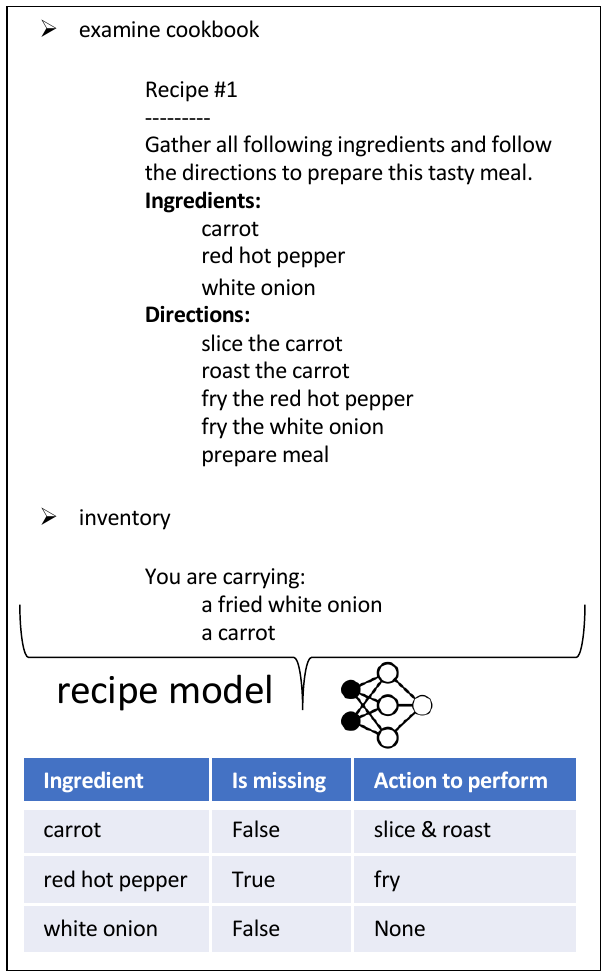}} \\
         (a) Based on the output from the neural recipe model, the inventory, and the room description the recipe manager constructs the mapping from high-level commands to parsable low-level commands. & 
         (b) The neural recipe model takes the raw strings of the recipe and the inventory as input and outputs a structured table specifying for every ingredient in the recipe if it still needs to be collected and which (cooking) action to perform with it. From this structured table the recipe manager constructs the parsable commands. \\
         \centering
         \includegraphics[width=.65\linewidth]{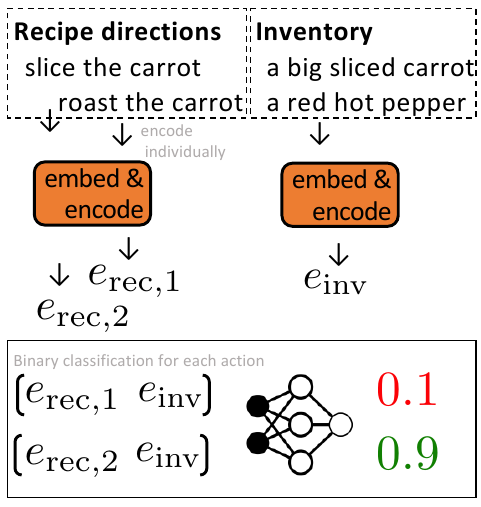} & \phantom{abc}\\
         (c) The binary \textit{Recipe Model} classifier that predicts for each recipe direction if it needs to be performed by the agent. & \phantom{abc} \\
    \end{tabular}
    \caption{Illustration of the recipe manager.}
    \label{fig:recipe_manager}
\end{figure}

\newpage

\section{Baseline Model Details}
For the two baseline models, LSTM-DQN and DRRN, we did a grid-search over a reasonable set of parameters. The best performing parameters are reported in Table \ref{tab:model_parameters} and used throughout the experiments.

\begin{table}[h]
    \centering
    \ra{1.1}
    \begin{tabular}{@{}rrr@{}}\toprule
    & LSTM-DQN & DRRN \\
    \midrule
    \textbf{Training parameters} \phantom{m}\\
     \cmidrule{0-0}
    Optimizer & Adam & Adam \\
    Learning rate & 0.001 & 0.005 \\
    Replay batch-size & 32 & 32 \\
    Discount factor & 0.5 & 0.9 \\
    \textbf{Model parameters} \phantom{mm} \\
    \cmidrule{0-0} 
    Embedding size & 64 & 100 \\
    Encoder hidden size & 192 & 256 \\
    MLP hidden size & 128 & 256 \\

    \bottomrule

    \end{tabular}
    \caption{Parameters of the best baseline models.}
    \label{tab:model_parameters}
\end{table}

\end{document}